\documentclass[11pt,a4paper]{article}
\usepackage[hyperref]{emnlp2018}
\usepackage{times}
\usepackage{latexsym}
\usepackage{amsmath}
\usepackage{url}
\usepackage{multirow}
\usepackage{graphicx}

\usepackage[ruled]{algorithm2e}
\usepackage[T2A,OT1]{fontenc}
\usepackage[utf8]{inputenc}

\usepackage{url}

\aclfinalcopy 

\newcommand{\pluseq}{\mathrel{+}=}

\title{A neural interlingua for multilingual machine translation}

\author{
	Yichao Lu\thanks{\quad Equal contribution} , Phillip Keung\footnotemark[1] , Faisal Ladhak, Vikas Bhardwaj, Shaonan Zhang, Jason Sun \\
	\tt \{yichaolu,keung,faisall,vikab,shaonanz,jasun\}@amazon.com \\
	Amazon Inc.
}

\date{}

\begin{document}
	\maketitle
	\begin{abstract}
		We incorporate an explicit neural interlingua into a multilingual encoder-decoder neural machine translation (NMT) architecture. We demonstrate that our model learns a language-independent representation by performing direct zero-shot translation (without using pivot translation), and by using the source sentence embeddings to create an English Yelp review classifier that, through the mediation of the neural interlingua, can also classify French and German reviews. Furthermore, we show that, despite using a smaller number of parameters than a pairwise collection of bilingual NMT models, our approach produces comparable BLEU scores for each language pair in WMT15.
		
	\end{abstract}
	
	\section{Introduction}
	\subsection{Multilingual Machine Translation}
	
	Neural machine translation (NMT) relies on word and sentence embeddings to encode the semantic information needed for translation. The standard attentional encoder-decoder models \cite{attention} for bilingual NMT decompose naturally into separate encoder and decoder subnetworks for the source and target languages. This factorization has inspired various forms of multilingual NMT models that extended the original bilingual framework to handle more language pairs simultaneously. We refer to NMT models that accept sentences from one source language and produce outputs in one target language as `bilingual'. We contrast this with `multilingual' NMT models, which support more than one source and/or target languages within the same model.
	
	The naive approach to multilingual machine translation would train a model for each language pair, which scales quadratically with the number of languages in the corpus. Instead, by combining language-specific encoders and decoders in different ways, \newcite{one_to_many}, \newcite{many_to_one}, \newcite{luong_multi}, and \newcite{firat_multi} have explored the one source-to-many target, many source-to-one target, and many source-to-many target multilingual MT settings. The multi-way shared attention model \cite{firat_multi} is closest to our work, in that they consider the large-scale, many-to-many scenario with multiple encoders and decoders.
	
	It is also possible to adapt existing bilingual NMT models to the many-to-many case without changing the architecture at all. The universal encoder-decoder approach \cite{universal_enc_dec, google} constructs a shared vocabulary for all languages in the dataset, and use just one encoder and decoder for multilingual translation. In addition, \newcite{google} introduce \emph{direct zero-shot translation}, which refers to the task of translating between language pairs without parallel text or pivoting through an intermediate language like English. Direct zero-shot translation may yield lower BLEU scores than pivot-based approaches, but avoids doubling the latency and computational overhead (due to translating the source sentence twice,) which is a concern for large-scale, productionized MT systems.
	
	Nonetheless, both the multi-way shared attention model and the universal encoder-decoder model suffer from certain disadvantages. For the former, direct zero-shot translation was shown to be impossible in \newcite{firat_zero}, and there is no indication that the model learns any kind of shared representation across languages. For the latter, the output vocabulary size is typically fixed to the vocabulary size for a single target language (i.e. roughly 20,000 to 30,000 types), regardless of the number of languages in the corpus. Increasing the vocabulary size is costly, since the training and inference time scales linearly with the size of the decoder's output layer.
	
	\subsection{Our Contributions}
	
	In this work, we construct an explicit \textit{neural interlingua} for multilingual NMT, which addresses some of the limitations in existing approaches. Our contributions are threefold:
	
	Firstly, we describe an attentional neural interlingua that receives language-specific encoder embeddings and produces output embeddings which are agnostic to the source and target languages.
	
	Secondly, we perform zero-shot translation (without pivot translation) for the Fr$\leftrightarrow$Ru, Zh$\leftrightarrow$Es and Es$\leftrightarrow$Fr pairs of the updated UN Parallel Corpus \cite{un}. At the time of writing, our approach is the only alternative to the universal encoder-decoder model for direct neural zero-shot translation. We observe a significant improvement in zero-shot translation performance compared to that model.
	
	Finally, we demonstrate that our model generates useful representations for crosslingual transfer learning. We use the source sentence embeddings from our translation model to create an English Yelp review classifier that can, through the mediation of the interlingua, classify French and German Yelp reviews. We also show that the sentence embeddings of parallel translations are close to each other in a low-dimensional space.

	\section{Model Architecture}
	
	\begin{figure}[h]
		\centering
		\includegraphics[scale=.24]{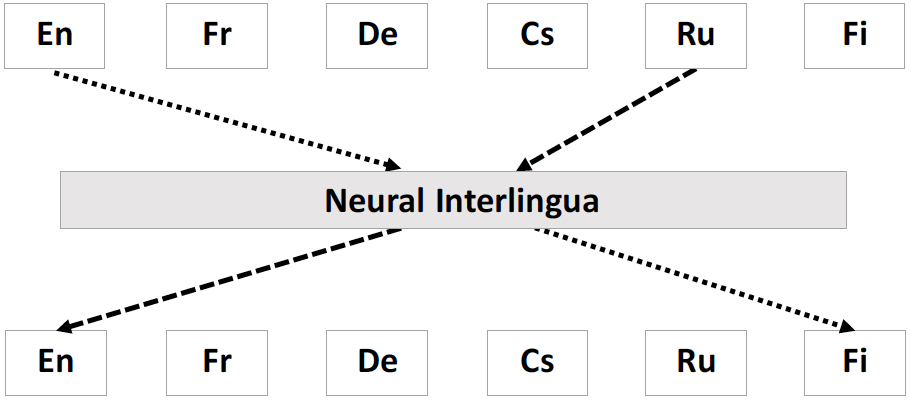}
		\caption{Our encoder-decoder model with the neural interlingua, trained on WMT15. The neural interlingua is an attentional encoder that converts language-specific embeddings to language-independent ones. Here, we illustrate the flow of data from English $\rightarrow$ Interlingua $\rightarrow$ Finnish, and Russian $\rightarrow$ Interlingua $\rightarrow$ English.}
		\label{interlingua_diagram}
	\end{figure}
	
	\begin{figure}[h]
		\centering
		\includegraphics[scale=.45]{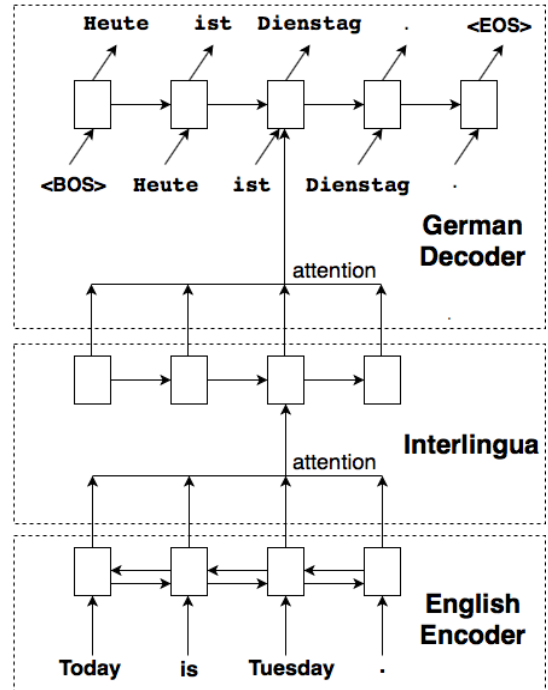}
		\caption{An in-depth look at the network structure when training/predicting with an En-De batch. The English sentence is fed through the English bidirectional LSTM encoder. The encoder states are passed into the neural interlingua, which is an attentional LSTM encoder. Finally, the hidden states of the interlingua are consumed by the German attentional LSTM decoder to generate the German translation.}
		\label{detailed_diagram}
	\end{figure}
	
	Figure \ref{interlingua_diagram} illustrates our basic model architecture. Each language has its own recurrent encoder and decoder. We attempt to construct a neural interlingua by passing the language-specific encoder embeddings through a shared recurrent layer, whose output embeddings are then passed to language-specific decoders.
	
	The figure describes the flow of data in the model; each minibatch only contains one source language and one target language, and only the parameters in the source encoder, interlingua, and target decoder are used for the forward and backward passes. During training, the source and target languages in each minibatch rotate according to a schedule (see Algorithm \ref{train_algo}). In Figure \ref{detailed_diagram}, we illustrate how an English sentence is converted into a German one.
	
	As with most sequence-to-sequence models, we can view the generation of the next token in the target sentence as the application of a series of neural network operations on the source sentence and the partial output thus far. We model the probability of each target sentence as follows,
	
	\begin{align*}
	p(y_i|y_{<i}, x) &= \text{Dec}_t(\text{Inter}(\text{Enc}_s(\text{Emb}_s(x))), \\
	&\qquad y_{i-1}, h^t_{i-1}) 
	\end{align*}
	
	where $y$ is the target sentence, $x$ is the source sentence, $\text{Dec}_t$ is the decoder for the target language $t$, $\text{Inter}$ is the neural interlingua, $\text{Enc}_s$ is the encoder for the source language $s$, $\text{Emb}_s$ is the word embedding matrix for $s$, $h^t_{i-1}$ is the state of the decoder at step $i-1$, $s\in\{1,...,S\}$ is the index of the source language, and $t\in\{1,...,T\}$ is the index of the target language.
	
	The source sentence $x$ is transformed from a sequence of one-hot representations to a sequence of word embeddings $B^s$ through $\text{Emb}_s$,
	
	\[ B^s = \text{Emb}_s(x^s) \]
	
	$B^s$ is a $b^s \times L_x$ matrix, where $L_x$ is the length of the source sentence, and $b^s$ is the size of the word embedding for the source language $s$.
	
	The sequence of word embeddings is converted into a sentence representation $E^s$ by $\text{Enc}_s$,
	
	\begin{align*}
	E^s_{.,i} &= \text{Enc}_s(B^s)_{.,i} \\
	&= \text{BiLSTM}(B^s_{.,i}, h^s_{i-1})
	\end{align*}
	
	$E^s$ is a $e^s \times L_x$ matrix, where $e^s$ is the size of encoder's output. The notation $X_{.,i}$ refers to the $i^\text{th}$ column of the matrix $X$. $\text{BiLSTM}$ is a bidirectional LSTM network, with forward and backward states $h^s_{i-1} = [\overrightarrow{h}^s_{i-1}, \overleftarrow{h}^s_{i+1}]$ for step $i-1$.
	
	The neural interlingua Inter is an attentional encoder that maps the language-specific representation $E^s$ to an interlingual representation $I$,
	
	\begin{align*}
	I_{.,i} &= \text{Inter}(E^s)_{.,i} \\
	&= W^I \lbrack \text{LSTM}(c^I_i , h_{i-1}^I), c_i^I \rbrack +b^I\\
	&= W^I \lbrack  h_i^I, c_i^I \rbrack +b^I
	\end{align*}
	
	where $h^I_{i-1}$ is the interlingua LSTM state for step $i-1$, $c^I_i = \sum_{j=1}^{L_x} \alpha^I_{ij} E^s_{.,j}$ is the attentional context vector, $\alpha^I_{ij} = \frac{exp(e^I_{ij})}{\sum_j exp(e^I_{ij})}$ and $e^I_{ij} = \text{MLP}_I(h_i^I, E^s_{.,j})$ are the normalized and unnormalized attention weights introduced in \newcite{attention}, and $z=[ x,y ]$ denotes the concatenation of the vectors $x$ and $y$ into a new vector $z$. We perform an affine transformation with $W^I, b^I$ to project the interlingua output to the desired dimensions.
	
	$I$ is a $e^i \times L_i$ matrix, where $e^i$ is the size of the interlingua's output. The output of the neural interlingua is always fixed in length to $L_i$ (where $L_i=50$ in our experiments), regardless of the length of the source sentence. We chose $L_i=50$ because, during model training, we restrict the maximum source sentence length to 50. To avoid learning language-specific embeddings, we do not use indicator tokens for the source or target languages.
	
	Finally, the decoder takes the interlingual representation $I$ and the partial target sentence $y_{<i}$ and computes the probability distribution for the next output token,
	
	\begin{align*}
	& p(y_i|y_{<i}, x) \\
	& = \text{Dec}_t(I, y_{i-1}, h^t_{i-1})_{.,i} \\
	& = \text{softmax}(W^t \lbrack \text{LSTM}(\lbrack y_{i-1}, c^t_i \rbrack, h^t_{i-1}),c^t_i \rbrack +b^t) \\
	& = \text{softmax}(W^t \lbrack h^t_i,c^t_i \rbrack +b^t) 
	\end{align*}
	
	where $c^t_i = \sum_{j=1}^{L_i} \alpha^t_{ij} I_{.,j}$ is the context vector at step $i$, and $\alpha^t_{ij}$ are the normalized attention weights. The decoders receive the source sentence only through the interlingual embedding.
	
	Like \newcite{firat_multi}, the number of encoders and decoders for our model architecture scales linearly (rather than quadratically) with the number of languages. In addition, since the neural interlingua provides a common source sentence representation to all decoders, the number of attention mechanisms also scales linearly with the number of languages.
	
	We note that the concept of a neural interlingua is independent of the architecture that is chosen. While we use a LSTM encoder-decoder model with single-headed attention for experimental simplicity, one could also introduce a neural interlingua to a transformer network \cite{transformer} or a CNN encoder-decoder network \cite{cnn} instead.
	
	\section{Experiments}
	
	We conducted 4 experiments with our model. 
	
	We compared the performance of bilingual NMT baselines against our proposed multilingual model, and observe comparable performance across all the language pairs in WMT15.
	
	We found that the language-independent sentence embeddings can be used for zero-shot multilingual classification. We train an English Yelp review classifier with the interlingual embeddings as input features, and use that model to classify French and German reviews.
	
	We performed direct zero-shot translation for 3 language pairs in the new UN Parallel Corpus. For this task, our model showed an improvement over the model architecture described in \newcite{google}. Our positive experimental finding confirms that our model provides a new approach for direct neural zero-shot translation.
	
	Finally, we visualized the language-independent sentence embeddings by projecting them down to 2 dimensions. We observe that parallel translations of French, German and English sentences remain close to each other in this low-dimensional space.
	
	\subsection{Model Training}
	
	\begin{table}[t!]
		\begin{center}
			\begin{tabular}{|c|c|c|}
				\hline \bf \multirow{2}{*}{Parameter} & \bf Multi & \bf \multirow{2}{*}{Bilingual}  \\ 
				& \bf -lingual &  \\ \hline
				vocabulary size & 30,000 & 30,000\\
				source embedding size &256&256\\
				target embedding size &256& 256\\
				output dimension & 512 & 512 \\
				encoder hidden size  & 512 & 512 \\
				decoder hidden size & 512 & 512 \\
				interlingua hidden size &512 & - \\
				interlingua length & 50 & - \\
				encoder depth & 2 & 4 \\
				interlingua depth & 1 & 0 \\
				decoder depth & 1 & 1\\
				attention type & additive & additive \\
				optimizer & Adam & Adam\\
				learning rate & 0.0002 & 0.0002\\
				batch size & 400 & 400 \\
				\hline
			\end{tabular}
		\end{center}
		\caption{Hyperparameters for the multilingual and bilingual encoder-decoder models.}
		\label{hps}
	\end{table}
	
	The hyperparameters for the bilingual baseline models and our multilingual network are summarized in Table \ref{hps}. Our multilingual model uses 1 bidirectional LSTM layer in the encoder for each input language, 1 attentional LSTM layer for the interlingua and 1 attentional LSTM layer in the decoder for each output language. The baseline bilingual models use 2 bidirectional LSTM layers in the encoder and 1 attentional LSTM layer in the decoder. We chose the Adam optimizer \cite{adam}, and we used importance sampling, as described in \newcite{sampled_softmax}, to accelerate model training.
	
	\subsection{Language Rotation During Training}
	
	\begin{algorithm} \label{train_algo}
		$\theta \leftarrow$ RandomInitializer() \\
		$schedule \leftarrow \{\}$ \\
		\For{S $\in$ \{En, Fr, De, Cs, Fi, Ru\}}
		{
			\For{L $\in$ \{Fr, De, Cs, Fi, Ru\}}
			{
				$schedule \pluseq \{(En, L), (L, En)\}$ \\
			}
			$schedule \pluseq \{(S, S)\}$ \\
		}
		\While{True}{
			\For{(s, t) $\in$ schedule}
			{
				$x_s \leftarrow SampleSource(s)$\\
				$y_t \leftarrow SampleTarget(t)$\\
				$a \leftarrow ForwardStep(\theta, x_s, y_t)$ \\
				$\nabla\theta \leftarrow BackwardStep(a, \theta)$ \\
				$\theta\leftarrow SGDUpdate(\theta, \nabla\theta)$
			}
		}
		\caption{Multilingual model training schedule on WMT15. We store the cycle of language pairs in $schedule$, and $x_s$ and $y_t$ refer to the source and target sentences respectively.}
	\end{algorithm}
	
	The language pair schedule used during training is crucial for learning an effective sentence representation. We provide the details in Algorithm \ref{train_algo}. In our initial experiments, we cycled through 10 language pairs (i.e. ($x$ $\rightarrow$ En, En $\rightarrow$ $x$), $x\in\text{\{Fr, De, Ru, Cs, Fi\}}$), where each minibatch consisted of sentences from one language pair. However, we found that the naive schedule failed to produce a useful representation for zero-shot translation or crosslingual text classification. Since WMT15 is not a multi-parallel corpus, the model essentially learns to handle two separate tasks, namely translation from English and translation to English. For instance, since the output of the De encoder and the En encoder would never be used by the same decoder, there is no reason for De and En source sentences to share the same embedding, even if they are translations of each other.
	
	To encourage the model to share the encoder representations across English and non-English source sentences, we added an extra identity language pair (i.e. De $\rightarrow$ De, En $\rightarrow$ En, etc.) to the rotation. The identity pair forces the source embeddings to be compatible with an additional decoder. We found that when we did not include the identity mapping task during training, the zero-shot BLEU score was $<1.0$ for the Fr-Ru language pair.

	\subsection{Multilingual NMT versus Bilingual NMT} \label{multi_exp}

\begin{table}[t!]
	\begin{center}
		\begin{tabular}{|c|c|c|c|}
			\hline \bf Source & \bf Target & \bf Bilingual  & \bf Multilingual \\ \hline
			\multirow{5}{*} {En} & Fr & \bf 34.85 & 33.80 \\
			& De & \bf 23.67 & 23.37 \\
			& Cs & \bf 17.60 & 16.62 \\
			& Ru & 21.26 & \bf 21.92 \\
			& Fi & 11.55 & \bf 13.34 \\
			\hline
			Fr & \multirow{5}{*} {En} & \bf 30.72 & 30.24 \\
			De &  & 27.08 & \bf 27.29 \\
			Cs &  & 23.00 & \bf 23.87 \\
			Ru &  & 24.14 & \bf 26.15 \\
			Fi &  & 14.77 & \bf 16.58 \\
			\hline
		\end{tabular}
	\end{center}
	\caption{Comparison of BLEU scores across language pairs in newstest2015 and newsdiscuss2015. We show the results for the bilingual baseline NMT models and our multilingual NMT model.}
	\label{bi_v_multi}
\end{table}

\begin{table*}[ht]
	\small
	\centering
	\begin{tabular}{|c|c|p{13cm}|}
		\hline \bf Color & \bf Lang. & \bf Text \\ \hline
		Green & En & spreads between sovereign bonds in Germany and those in other countries were relatively unaffected by political and market uncertainties concerning Greece in late 2014 and early 2015 . \\
		& Fr & par contre , la différence entre les obligations souveraines allemandes et celles d'autres pays a été relativement peu touchée par les incertitudes politiques et les doutes des marchés concernant la Grèce fin 2014 et début 2015 . \\
		& Ru & {\fontencoding{T2A}\selectfont 
			политическая и рыночная нестабильность , связанная с ситуацией в Греции в конце 2014 - го и начале 2015 года , практически не отразилась на спредах доходности между государственными облигациями Германии и других стран . } \\ \hline
		Red & En & 13 . we underscore the need to accelerate efforts at all levels to achieve the objectives of the international arrangement on forests beyond 2015 and the need to establish a stronger , more effective and solid arrangement for the period 2015 to 2030 ; \\
		& Fr & 13 . nous soulignons qu’il faudra redoubler d’efforts à tous les niveaux pour atteindre les objectifs de l’arrangement international après 2015 et qu’il faudra mettre en place un arrangement plus solide et plus efficace pour la période 2015 - 2030 ; \\
		& Ru & {\fontencoding{T2A}\selectfont 13 . мы подчеркиваем , что необходимо активизировать усилия на всех уровнях в интересах достижения целей международного механизма по лесам на период после 2015 года и создать действенный , более эффективный и надежный механизм на период 2015 - 2030 годов ;} \\ \hline
		Orange & En & the various training activities are listed in table 2 below . \\
		& Fr & on énumère dans le tableau 2 ci - dessous les diverses activités de formation . \\
		& Ru & {\fontencoding{T2A}\selectfont в представленной далее таблице 2 приведен перечень различных мероприятий по профессиональной подготовке .} \\ \hline
		Blue & En & the Conference affirms that , pending the realization of this objective , it is in the interest of the very survival of humanity that nuclear weapons never be used again . \\
		& Fr & elle affirme que , en attendant la réalisation de cet objectif , il est dans l’intérêt de la survie même de l’humanité que les armes nucléaires ne soient plus jamais utilisées . \\
		& Ru & {\fontencoding{T2A}\selectfont 
			конференция заявляет , что , пока эта цель не достигнута , необходимо в интересах самого выживания человечества добиться того , чтобы ядерное оружие никогда не было вновь применено .} \\ 
		\hline
	\end{tabular}
	\caption{Text of the parallel sentences in Figure \ref{interlingua_embeddings}.}
	\label{sentences}
\end{table*}

We used the training corpora from the WMT15 translation task to train our encoder-decoder models. The dataset provides English $\leftrightarrow$ (German, French, Czech, Russian, Finnish) parallel sentences. We followed the standard WMT preprocessing recipes\footnote{e.g. \url{http://data.statmt.org/wmt17/translation-task/preprocessed/de-en/prepare.sh}}, which are based on the Moses library \cite{moses}. For each language, we created a vocabulary of 30,000 word pieces using byte pair encoding \cite{bpe}. Sentences longer than 50 word pieces were removed from the training corpus. We used newstest2014 and newsdev2015 as our development set, and newstest2015 and newsdiscuss2015 as our test set.

We compared the performance of the multilingual model against bilingual baseline models. The BLEU scores are provided in Table \ref{bi_v_multi}. Results are reported on newstest2015 and newsdiscuss2015. We see that, while the performance is broadly similar (i.e. generally $\textless 1.0$ BLEU) between the our model and the baselines, there is a decrease in BLEU for higher-resource languages (e.g. Fr) and an increase in BLEU for lower-resource languages (e.g. Fi, Ru). We suspect that this is a consequence of the language pair schedule, which cycles through all language pairs as though they were equally frequent in the corpus. A similar effect was also observed in \newcite{google}.

\newcite{copied_corpus} have shown that (specifically in low-resource settings) using copied monolingual data can improve model performance. We followed the technique in \newcite{copied_corpus} to strengthen the baseline models, but did not observe an improvement in the final BLEU score. This may be due to the fact that even the smallest language pair in WMT15 has 2 million sentence pairs, which is more than 3 times larger than either the Tr-En or Ro-En pairs discussed in \newcite{copied_corpus}.

As with \newcite{firat_multi}, we generally see an improvement when translating to English. We believe that this is because the English language model is stronger in the multilingual case, since the English decoder sees more English text.

\subsection{Zero-shot Multilingual Classification}

\begin{table*} 
	\begin{center}
		\begin{tabular}{|c|c|c|c|}
			\hline   & \multicolumn{3}{c|}{\bf Input Language} \\ \cline{2-4}
			\bf  & En & De & Fr \\ \hline
			\bf Trigram & 91.6\% $\pm$ 0.9\% & 89.6\% $\pm$ 0.9\% & 91.5\% $\pm$ 0.9\% \\ \hline
			\bf Embeddings & 91.5\% $\pm$ 0.9\% & 89.2\% $\pm$ 0.9\% & 91.1\% $\pm$ 0.9\% \\ \hline
			\bf \% Positive & 82.9\% & 86.7\% & 88.5\% \\ \hline
		\end{tabular}
	\end{center}
	\caption{Accuracy for crosslingual Yelp binary review classification. The trigram baseline model was trained on English reviews, and tested on English reviews and English translations of French and German reviews. The embedding-based classifier uses interlingual embeddings from our model in Section \ref{multi_exp}. `\% Positive' refers to the proportion of the test set that has a positive label.}
	\label{yelp}
\end{table*}

We constructed a multilingual Yelp review dataset from a subset of the Yelp Challenge (Round 10) corpus. We restrict ourselves to English, French, and German reviews. The training corpus consists of 5,000 English Yelp reviews, and the test sets contain 4,000 reviews for each language. The French and German reviews were extracted by applying language detection on reviews from Quebec, Canada and Baden-W\"urttemberg, Germany. The review scores were binarized, where 4 and 5 star reviews were labeled as positive, and 1 and 2 star reviews were labeled as negative. We reuse the encoders trained in Section \ref{multi_exp} in this section's experiments.

At training time, an English Yelp review is treated as one sentence; we do not apply sentence segmentation to the review. It is passed through the English encoder, and the neural interlingua converts the English sentence representation to a fixed-length representation. To create a feature vector for the text classifier, we apply mean-pooling to the sentence representation. Under our experimental settings, every sentence is converted to a $512 \times 50$ interlingual embedding, which is mean-pooled into a 512-dimensional vector. We then fit a logistic regression model using this feature vector and the sentence polarity as the binary label. The classifier is only trained on English reviews.

At prediction time, we pass the text of a German review through the German encoder and the interlingua, which is again mean-pooled to form a 512-dimensional vector. Since the interlingual representation should be language-independent, we can attempt to classify German reviews by providing the vector representation of the German review to the English classifier. We use the same process for French reviews.

In Table \ref{yelp}, we compare the accuracy of the classifier trained on English review embeddings to that of a baseline model. We established the baseline by training a trigram classifier on the English reviews, and used English translations of the French and German reviews for classification. We obtained the translations through the Google Translate API. The classification accuracy using the interlingual embeddings or the translated French and German reviews are similar, which shows that the embeddings have retained semantic information in a language-independent way.

	\subsection{Direct Zero-shot Translation}
	
	\begin{table*} 
		\begin{center}
			\begin{tabular}{|l|r|r|r|r|r|r|}
				\hline & \bf Fr-Ru & \bf Ru-Fr & \bf Es-Zh & \bf Zh-Es & \bf Es-Fr & \bf Fr-Es \\ 
				\hline
				\bf This Work & 18.24 & 21.61 & 17.66 & 18.66 & 30.08 & 31.94 \\
				\hline
				\bf Univ. Enc-Dec & 8.77 & 9.76 & 8.62 & 6.13 & 15.04 & 14.37 \\
				\hline
				\bf Pivot  & 20.87 & 27.34 & 26.03 & 26.01 & 31.84 & 32.93 \\
				\hline
				\bf Direct NMT  & 28.29 & 33.26 & 32.36 & 32.69 & 41.38 & 44.49 \\
				\hline
			\end{tabular}
		\end{center}
		\caption{Zero-shot BLEU scores on the UN Parallel Corpus on selected language pairs. The universal encoder-decoder, pivot and direct NMT results were retrieved from \newcite{tree_pivot}. Our proposed model outperforms the universal encoder-decoder model \cite{google} on the zero-shot translation task.}
		\label{zero_shot_exp}
	\end{table*}
	
	The updated UN Parallel Corpus \cite{un}, unlike the WMT corpus, is a fully multi-parallel corpus that contains English, Spanish, French, Arabic, Chinese and Russian text. We used this corpus as a testbed for our zero-shot translation experiments.
	
	We trained our multilingual model on the UN corpus, following the same settings that we used for the WMT corpus (see Table \ref{hps} and Algorithm \ref{train_algo}). The text was processed following the steps provided in \newcite{tree_pivot}. We restrict the training corpus to sentence pairs that have English as either the source or target language. 
	
	We used the Fr-Ru, Es-Zh and Es-Fr portions of the test set from the UN corpus for the zero-shot translation evaluation. The training dataset that we constructed does not contain direct Fr-Ru, Es-Zh or Es-Fr sentence pairs. The test set contains 4,000 sentence pairs for each language pair.
	
	We examine the BLEU scores for zero-shot translation on the UN corpus in Table \ref{zero_shot_exp}. The universal encoder-decoder, pivot and direct NMT results were retrieved from \cite{tree_pivot}. By `direct NMT', we refer to a model trained directly on the parallel text.
	
	Our multilingual model performs significantly better on the direct zero-shot task than the universal encoder-decoder approach of \newcite{google}. Generally, our model does not perform as well as the pivot approach, though in the case of Es-Fr and Fr-Es, the difference is surprisingly small (\textless 2.0 BLEU).
	
	Improving direct zero-shot methods to reach parity with pivot translation has practical consequences for large-scale NMT systems, like reduced latency and computational overhead. (Recall that pivot translation must translate every source sentence twice; first into the intermediate language, and then into the target language.) Our results show progress towards the goal of transitioning away from pivot-based methods to neural zero-shot translation.
	
	\subsection{Interlingua Visualization}
	
	\begin{figure}[h]
		\centering
		\includegraphics[scale=.55]{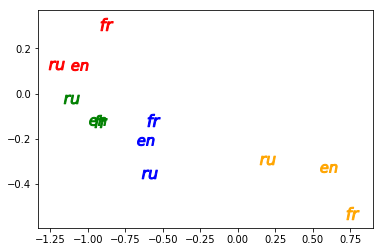}
		\caption{Interlingual embeddings for four groups of parallel English, French, and Russian sentences from the UN Parallel Corpus. The 512-dimensional mean-pooled interlingual sentence embeddings were projected down to $\textbf{R}^2$ using PCA. Refer to Table \ref{sentences} for the colors and text of the sentences.}
		\label{interlingua_embeddings}
	\end{figure}
	
	In Figure \ref{interlingua_embeddings}, we plot the embeddings for 4 groups of parallel sentences. Sentences from the same group share the same color. Each group contains one French, one English and one Russian sentence which are parallel translations of each other. We provide the text of the embedded sentences in Table \ref{sentences}.
	
	The embeddings were generated by mean-pooling each sentence embedding to a 512-dimensional vector and projecting it to $\textbf{R}^2$ using PCA. From the figure, we observe a clear separation between different groups of sentences, while sentences within the same group remain close to each other in space. This is the expected outcome if our model has captured language-independent semantic information in its sentence representations.
	
	\section{Related Work}
	
	\subsection{Networks with Language-specific Encoders and Decoders}
	
	The many-to-one approach explored in \newcite{many_to_one} primarily considers the trilingual case, where a multi-parallel corpus is available, and uses 2 encoders simultaneously to provide the source context for the decoder. We note that using 2 encoders simultaneously requires having 2 source sentences for every desired target sentence at prediction time, which is not the setting that we investigate here.
	
	By combining a single encoder with multiple attentional decoders, the one-to-many approach presented in \newcite{one_to_many} showed an improvement in translation performance, due to the increase in the number of sentences seen by the encoder and through multi-task learning.
	
	The many-to-many approach in the shared attention model \cite{firat_multi} assigns a different encoder and decoder to each language, but shares the decoders' attention mechanisms. By specifying a `universal' attention mechanism for all language pairs, \newcite{firat_multi} avoid creating as many attention mechanisms as there are language pairs (i.e. avoids quadratic scaling).
	
	However, the attention mechanism acts as the alignment model between the source and target sentences, and a shared attention mechanism may be too restrictive, especially for languages that have very different word orders. Our interlingual approach relaxes the requirement of a single, shared attention mechanism. In our framework, there are as many attention mechanisms as there are decoders.
	
	\subsection{Universal Encoder-Decoder Networks}
	
	\newcite{google} have foregone the use of multiple encoders and decoders, and instead use one universal encoder and one universal decoder. They constructed a joint vocabulary for all languages in the corpus, consisting of word pieces derived from a byte-pair encoding \cite{bpe} on the union of the vocabulary of all the languages, and include special tokens to indicate what the output language should be. \newcite{universal_enc_dec} follow a similar approach, but the shared vocabulary is constructed by prepending a language identifier to each token.
	
	The universal encoder-decoder approach does have some shortcomings. \newcite{google} rely on the existence of a shared vocabulary, which may not be as sensible in some combinations (e.g. Chinese and English) as in others (e.g. Spanish and Portuguese). If the languages' vocabularies do not share many word pieces, then either the decoder's output layer will be very large, which slows down training and inference, or the output layer will be artificially constrained to a manageable size, which impacts translation performance.
	
	Our approach, on the other hand, allows each target language to retain its own decoder. The total vocabulary size can then expand with the number of languages without affecting training or inference speed.
	
	\subsection{Zero-shot Translation}
	
	One of the challenges in multilingual MT is data sparsity, which refers to the lack of parallel text for every possible language pair in a corpus. Zero-shot translation is the task of translating between language pairs without parallel text.
	
	An early approach to allow zero-shot translation made use of a `pivot' language in the translation process \cite{pivot}. For instance, in sentence-based pivoting, the source sentence is translated into a pivot language, and from the pivot language translated to the target language. Various extensions of the pivot technique have been proposed over the years, see \newcite{pivot_comparison}, \newcite{factored_pivot}, \newcite{tree_pivot}, \newcite{cohn_triangulation}.
	
	Universal encoder-decoder systems like \newcite{google} have demonstrated the ability to perform direct zero-shot translation without using a pivot language at all, albeit with a significant BLEU reduction for some language pairs.
	
	\section{Conclusion}
	
	We incorporate a neural interlingua component into the standard encoder-decoder framework for multilingual neural machine translation, and demonstrate that the resulting model learns language-independent sentence representations, enabling zero-shot translation and crosslingual text classification.
	
	We perform direct zero-shot translation for 3 language pairs without pivoting through an intermediate language like English. We observe an improvement in zero-shot translation performance compared to the universal encoder-decoder results reported in \newcite{tree_pivot}. Furthermore, we use the learned encoder to train an English Yelp review classifier that can, with the help of the interlingual embeddings, also classify German and French reviews. Finally, our experiments showed that the results from our model are comparable to the results from bilingual baselines. 
	
	In future work, we intend to address the significant performance gap between direct neural zero-shot translation and pivot translation. By manipulating the sentence embeddings in an appropriate way, we aim to extract significant improvements over the results presented in this paper.
	
	
	
	\bibliography{emnlp2018}
	\bibliographystyle{acl_natbib_nourl}

	\appendix
	
\end{document}